\documentclass[a4paper,11pt]{article}

\usepackage{algorithm}
\usepackage{array}
\usepackage{amsmath}
\usepackage{amssymb}
\usepackage[utf8]{inputenc}
\usepackage[export]{adjustbox}
\usepackage{textcomp}
\usepackage{stfloats}
\usepackage{csquotes}
\usepackage[super]{nth}
\usepackage{caption}
\usepackage{url}
\usepackage{verbatim}
\usepackage{siunitx}
\usepackage[font={footnotesize}, textfont=sf]{subfig} 
\usepackage[table,xcdraw]{xcolor}
\usepackage{fourier}
\usepackage{color}
\usepackage{multirow}
 \usepackage{graphicx}
\usepackage{url}
\usepackage[affil-it]{authblk}
\usepackage{amsmath}
\usepackage{wrapfig}
\usepackage{booktabs}
\usepackage{threeparttable}
\usepackage[citebordercolor=white, linkbordercolor=white, urlbordercolor=black, hidelinks, pdfborderstyle={/S/U/W 1}, breaklinks]{hyperref}
\usepackage{cleveref}

\usepackage[T1]{fontenc}
\usepackage{times}

\begin{document}

\title{SS-SFR: Synthetic Scenes Spatial Frequency Response on Virtual KITTI and Degraded Automotive Simulations for Object Detection}

\author{Daniel Jakab$^{1,}$$^5$, Alexander Braun$^2$, Cathaoir Agnew$^1$, Reenu Mohandas$^{1,}$$^5$, Brian Michael Deegan$^{3,}$$^5$, Dara Molloy$^4$, Enda Ward$^4$, Anthony Scanlan$^1$, Ciarán Eising$^{1,}$$^5$}
\affil{$^1$Dept. of Electronic and Computer Engineering, University of Limerick, Castletroy, Co. Limerick  V94 T9PX, Ireland\\
$^2$Faculty of Electrical Engineering \& Information Technology , University of Applied Sciences, Düsseldorf 40476, Germany\\
$^3$Dept. of Electrical and Electronic Engineering, University of Galway, Galway H91 TK33, Ireland\\
$^4$Valeo Vision Systems, Tuam, Galway DY1 22DJ, Ireland\\
$^5$Lero the Science Foundation Ireland Research Centre for Software, University of Limerick, Limerick, V94 T9PX, Ireland\\
}
\date{}
\maketitle
\thispagestyle{empty}

\emph{This paper is a preprint of a paper submitted to the 26th Irish Machine Vision and Image Processing Conference (IMVIP 2024). If accepted, the copy of record will be available at IET Digital Library.}

\begin{abstract}
Automotive simulation can potentially compensate for a lack of training data in computer vision applications. However, there has been little to no image quality evaluation of automotive simulation and the impact of optical degradations on simulation is little explored. In this work, we investigate Virtual KITTI and the impact of applying variations of Gaussian blur on image sharpness. Furthermore, we consider object detection, a common computer vision application on three different state-of-the-art models, thus allowing us to characterize the relationship between object detection and sharpness. It was found that while image sharpness (MTF50) degrades from an average of 0.245cy/px to approximately 0.119cy/px; object detection performance stays largely robust within 0.58\%(Faster RCNN), 1.45\%(YOLOF) and 1.93\%(DETR) across all respective held-out test sets.
\end{abstract}

\section{Introduction}
Autonomous vehicles require large amounts of data to anticipate safety-critical scenarios and corner cases on the road. However, even the largest and most influential automotive datasets (such as BDD100k \cite{yu2020bdd100k} and KITTI \cite{geiger2013vision}) only capture a small subset of possible scenarios.
Simulation is promising as it is relatively feasible to set up an environment that can reflect real life and be utilized for perception systems. The main issue with simulation is that there has been little to no quality evaluation of this data. There are considerable differences between simulation and real life as discussed in a recent survey \cite{jakab2024surround} where automotive simulators lack the simulation of lenses from the real world.
We can refer to this concept as the 'syn-to-real-gap' where the need to bridge the gap between simulation and real life must be realised. As a means of measuring the image quality of automotive simulation, we measure the Modulation Transfer Function(MTF) curve where it falls off to 50\%, also known as MTF50 and is a typical measurement for cameras adopted by the industry \cite{mueller2020simulating, burns2022updated}. MTF50 is a scalar value selected from the MTF curve which indicates image sharpness in capturing details through the lens of a camera.
We use the sfrmat5 algorithm by Peter Burns with updates made recently~\cite{burns2022updated}, a Matlab software script function, which provides a Spatial Frequency Response (SFR) (otherwise known as the Modulation Transfer Function) from digital image files using the Slanted Edge Method. It follows the ISO12233:2023 standard recently published \cite{ISOBS2023} to resolve photographic camera measurements \cite{da2021slanted}. We obtain the measurement of MTF50 directly from the simulation scenes of Virtual KITTI. Recent work has shown that natural scenes can be measured in public datasets to obtain the Edge Spatial Frequency Response (e-SFR), especially in autonomous vehicles \cite{jakab2024measuring, van2021natural}. This strategy is also known as Natural Scenes Spatial Frequency Response (NS-SFR) \cite{van2023tool, van2022camera,van2021natural}.
In this paper, we apply the NS-SFR approach to synthetic scenes in simulation called Synthetic Scenes Spatial Frequency Response (SS-SFR) using the Virtual KITTI dataset. The main contributions of this paper are:
\begin{itemize}
    \item Assess simulation as an alternative to real-world data collection.
    \item Degrade simulation with Gaussian blur\cite{szeliskicomputer} to simulate the effects of an out-of-focus lens.
    \item Investigate how image sharpness is impacted in the degraded simulations.
    \item Compare image sharpness to object detection to see how computer vision behaves with degradation.
\end{itemize}

\section{Related Work}
\label{sec:rel-work}
Automotive simulation has recently emerged as a new field in the research of autonomous vehicles. Blueprint measurements of road test tracks for autonomous vehicles have been proven useful to reconstruct road surfaces and terrain in ASAM OpenDRIVE simulations\cite{ASAM2024}. Simulations of this nature are also widely known as model-driven simulations. As an example,
virtual simulation models were publicly released for ZalaZONE\cite{ZalaZONE2020}, an automotive proving ground for autonomous driving research. These models can be exported to any model-driven simulator (i.e. Unity, Unreal Engine, and Vires VTD, to name but a few).
The latest research shows a clear trend towards the use of simulation in computer vision models \cite{gaidon2016virtual, cabon2020virtual, saad2019camera, carlson2019modeling}.
The Virtual KITTI dataset, a model-driven simulation dataset created in the Unity Simulator, was first released by \cite{gaidon2016virtual}. This was one of the first automotive simulation datasets annotated for computer vision. Qualitatively, the dataset lacked fundamental aspects of photorealism but was sufficient for training algorithms. The visual quality of automotive simulation has drastically improved since then \cite{cabon2020virtual}. Despite this, considerable differences remain compared to real-life. A recent concept \cite{carlson2019modeling} shows some work in degrading automotive simulation using image augmentations to introduce realistic optical artefacts for computer vision. However, there is no evaluation of image quality in this work which suggests an inconclusive analysis. In \cite{mueller2020simulating}, a framework is demonstrated to simulate and validate the realism of camera simulations using the ISO12233 standard for camera measurement. ISO12233 is an established method to measure the resolution of a camera system also known as the SFR \cite{mueller2020simulating}. This framework shows how closely simulations can model the real-life lens from the laboratory \cite{von1934beugungstheorie}). Inspired by both sets of work \cite{carlson2019modeling, mueller2020simulating}, it is interesting to further investigate how the ISO12233 standard with its latest updates\cite{ISOBS2023} can validate automotive simulation.

\section{Methodology}
\begin{figure*}[t]
    \centering
    \vspace{-20pt}
    \captionsetup[subfloat]{captionskip=-70pt}
    \subfloat{\includegraphics[width=5.5in, keepaspectratio]{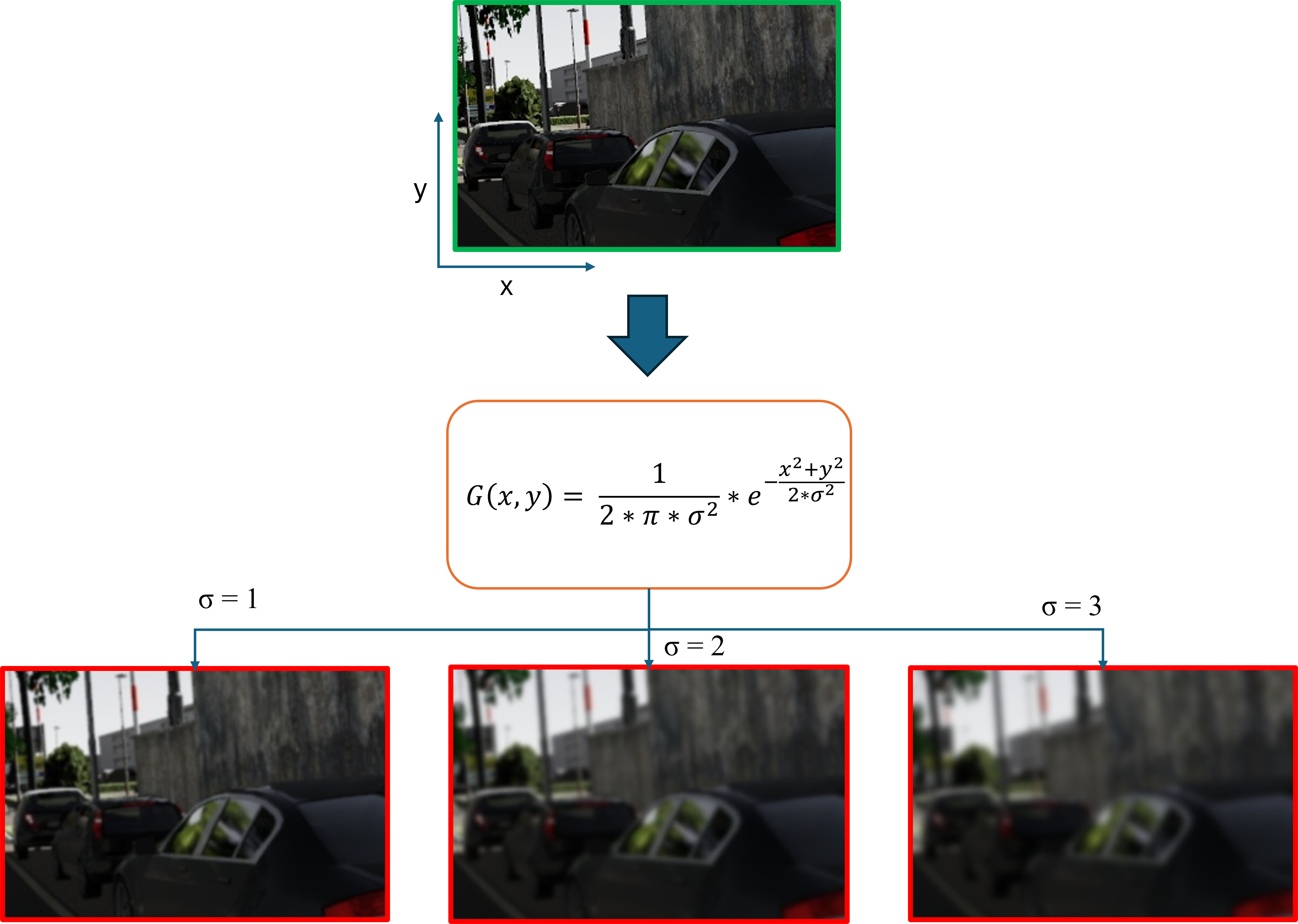}}
    \vspace{-10pt}
    \caption{Virtual KITTI Gaussian blur image degradation between $\sigma=[1,2,3]$.}
    \label{fig:sim-deg}
    \vspace{-0pt}
\end{figure*}
To understand how image sharpness behaves in simulation, we shall proceed to do the following:
\begin{enumerate}
    \item{Choose the clone camera perspectives from the latest version of Virtual KITTI, consisting of 2,126 images in total \cite{cabon2020virtual}. }
    \item{Perform degradations by using a Gaussian blur filter~\cite{szeliskicomputer}:
    \begin{equation}
        G(x,y) = \frac{1}{2*\pi*\sigma^2}*e^{-\frac{x^2 + y^2}{2*\sigma^2}} \label{eq:gauss}
    \end{equation}
    Where G(x,y) is the degraded two dimensional image in terms of x and y where $\sigma=[1,2,3]$ and kernel size is determined by:
    \begin{equation}
        k = 6*\sigma + 1 \label{eq:k_size}
    \end{equation}
    As shown in Figure \ref{fig:sim-deg}, Virtual KITTI is degraded in three iterations by adjusting the value of $\sigma=[1,2,3]$ in Equation \ref{eq:gauss}. Therefore, the kernel sizes for each $\sigma$ value are $(k_{\sigma},k_{\sigma}) = [(7,7),(13,13),\\(19,19)]$, respectively. This degradation approach uses the OpenCV library\cite{OpenCVGaussianBlur} to apply the Gaussian blur filter on each image creating a degraded variation of the original dataset.
    }
    \item{Apply NS-SFR as demonstrated in previous work \cite{jakab2024measuring,van2023tool} to isolate and extract regions of valid slanted edges from Virtual KITTI. All four dataset variations were processed using all scenes with the clone camera perspectives.}
    \item{Measure image sharpness using the ISO12233:2023 Slanted Edge Method (sfrmat5)~\cite{ISOBS2023,burns2022updated} on all extracted edge regions from the images of Virtual KITTI before and after each of the three Gaussian degradations. Results are averaged for each of the four dataset variations across all clone perspectives of Virtual KITTI.}
    \item{We utilize annotated Virtual KITTI for object detection where the dataset is split into 80\% training, 10\% validation, and 10\% test. The same split is used for all four variations of Virtual KITTI where each degradation split contains the same images.
    \item{Three different state-of-the-art object detection models are chosen for training where the $1242\times375$ resolution images are resized according to default configurations for the models \cite{mmdetection} with a batch size of 2 in each training run:
    \begin{enumerate}
        \item Faster RCNN\cite{Ren_2017}, images resized by default to a scale of $1333\times800$ (see default configuration~\cite{mmdetection}).
        \item YOLOF\cite{chen2021you}, images resized by default to $1333\times800$ (see default configuration\\~\cite{mmdetection}).
        \item DETR\cite{detr}, images resized by default between $1333\times480$ and $1333\times800$ (see training approach in the original work \cite{detr}).
    \end{enumerate}
    }
    All three models are trained on an NVIDIA GeForce RTX 3080. As a method for a controlled experiment, the same ResNet50\cite{he2016deep} backbone is used for all three networks.  The stopping criteria for both the Faster RCNN and YOLOF models are determined using `1x' configuration from the MMDetection toolbox \cite{mmdetection} which has a 12 epoch training cycle and the learning rate decays by a factor of 10 at the \nth{8} and \nth{11} epochs, respectively. For DETR, the default learning policy of 150 epochs is used where the learning rate decays by a factor of 10 at the 100th epoch. Results and training logs for 150 epochs are provided in the original work \cite{detr}. Transformer models typically take longer to converge whereas the YOLOF model can reach the same performance with seven times fewer training epochs than DETR \cite{chen2021you}.
    }
\end{enumerate}
\section{Results}
In this section, we present the SS-SFR of Virtual KITTI and object detection performance by comparing image sharpness(MTF50) to mean Average Precision(mAP).
\subsection{Virtual KITTI SS-SFR}
For SS-SFR measurements, the Virtual KITTI clone perspective was evaluated in four variations:
\begin{enumerate}
    \item Original or baseline dataset(i.e. No degradation applied).
    \item Degradation for $\sigma=1, [(k_{\sigma},k_{\sigma})[=[(7,7)]$.
    \item Degradation for $\sigma=2, [(k_{\sigma},k_{\sigma})[=[(13,13)]$.
    \item Degradation for $\sigma=3, [(k_{\sigma},k_{\sigma})[=[(19,19)]$.
\end{enumerate}
\subsubsection{Horizontal Edges}
Horizontal MTF50 (HMTF50) can be seen in Table \ref{tab:results}, four different results can be measured for MTF50. The baseline dataset was measured to be around 0.234cy/px. This is notably higher than KITTI measurements from previous work where horizontal edges averaged between 0.20-0.22cy/px \cite{jakab2024measuring}. This indicates that Virtual KITTI tends to have sharper edges than KITTI images taken from a real-life camera. 

Upon degrading the Virtual KITTI images notable image sharpness degradation can be measured. With $\sigma=1$ and kernel size of $(k_{1},k_{1}) = (7,7)$ MTF50 results degraded by 31.62\%. With $\sigma=2$ and kernel size of $(k_{2},k_{2}) = (13,13)$ MTF50 results degraded by 49.15\%. Finally, with $\sigma=3$ and kernel size of $(k_{3},k_{3}) = (19,19)$ MTF50 results degraded produced the same percentage as for the previous degradation(i.e. 49.15\%). Please see Table \ref{tab:results} for HMTF50 results. For horizontal edges, both $\sigma=2$ and $\sigma=3$ columns generated the largest degradations compared to the baseline indicating the little difference between both. The degradation of edges follows that of a typical Gaussian blur with a radius of 2 which can be measured around 0.117cy/px from a previous study\cite{Imatest2024}.

\begin{table}[t]
\centering
\footnotesize
\caption{Comparison of degraded Virtual KITTI MTF50 measurements for both horizontal and vertical edges to the baseline dataset.}
\label{tab:results}
\begin{tabular}{|l|l|ll|ll|ll|}
\hline
sigma    & kernel size & HMTF50 & \%                            & VMTF50 & \%                            & MTF50(mean) & \%                            \\ \hline
baseline & baseline    & 0.234  & -                             & 0.256  & -                             & 0.245       & -                             \\
1        & (7,7)       & 0.160  & {\color[HTML]{FE0000} -31.62} & 0.159  & {\color[HTML]{FE0000} -37.89} & 0.160       & {\color[HTML]{FE0000} -34.69} \\
2        & (13,13)     & 0.119  & {\color[HTML]{FE0000} -49.15} & 0.122  & {\color[HTML]{FE0000} -52.34} & 0.120       & {\color[HTML]{FE0000} -51.02} \\
3 &
  (19,19) &
  \textbf{0.119} &
  {\color[HTML]{FE0000} \textbf{-49.15}} &
  \textbf{0.119} &
  {\color[HTML]{FE0000} \textbf{-53.52}} &
  \textbf{0.119} &
  {\color[HTML]{FE0000} \textbf{-51.43}} \\ \hline
\end{tabular}%
\begin{tablenotes}
  \small
  \item {H/V}MTF50 = MTF50 for both horizontal(H) and vertical(V) edges where the mean of both is taken as \emph{MTF50(mean)}.
\end{tablenotes}
\end{table}

\begin{table}[t]
\centering
\caption{Object Detection Results on all four degradations of Virtual KITTI.}
\label{tab:cv-results}
\resizebox{\columnwidth}{!}{%
\begin{tabular}{lllllllll}
\hline
Model                        & sigma    & kernel size & mAP0.50:0.95(\%) & mAP0.5(\%) & mAP0.75(\%)    & $mAP0.5:0.95_{s}$(\%) & $mAP0.5:0.95_{m}$(\%) & $mAP0.5:0.95_{l}$(\%) \\ \hline
\multirow{4}{*}{Faster RCNN} & baseline & baseline    & \textbf{69.45}   & 81.24      & \textbf{76.46} & 51.28             & \textbf{86.76}     & 90.20              \\
 & 1 & (7,7)   & 69.05 & 80.54          & 75.57          & 51.13          & 86.31          & \textbf{90.68} \\
 & 2 & (13,13) & 69.06 & 80.59          & 75.22          & 51.02          & 86.27          & 90.20          \\
 & 3 & (19,19) & 69.17 & \textbf{81.75} & 75.16          & \textbf{56.09} & 85.56          & 90.02          \\ \hline
\multirow{4}{*}{YOLOF}       & baseline & baseline    & \textbf{60.06}   & 72.23      & 65.89          & 26.16             & 80.25              & \textbf{90.90}     \\
 & 1 & (7,7)   & 59.82 & 72.35          & 66.22          & 30.80          & \textbf{80.29} & 89.51          \\
 & 2 & (13,13) & 59.70 & \textbf{72.36} & 65.90          & \textbf{33.03} & 79.34          & 89.81          \\
 & 3 & (19,19) & 59.20 & 71.64          & \textbf{66.72} & 30.58          & 78.38          & 90.16          \\ \hline
\multirow{4}{*}{DETR}        & baseline & baseline    & 61.91            & \textbf{86.33}      & 65.90          & 43.95             & 72.95              & \textbf{91.09}              \\
 & 1 & (7,7)   & \textbf{62.28} & 85.82          & \textbf{67.19}          & 41.83          & \textbf{73.04}          & 90.41          \\
 & 2 & (13,13) & 61.15 & 84.49          & 63.79          & \textbf{46.18}          & 70.54          & 90.46          \\
 & 3 & (19,19) & 61.10 & 85.32          & 65.32          & 42.32          & 70.55          & 89.12          \\ \hline
\end{tabular}%
}
\begin{tablenotes}
  \small
  \item Note: all mAP values are recorded between an Intersection over Union(IoU) of 0.5:0.95 except for mAP0.5(\%) where IoU = 0.5 and mAP(0.75) where IoU = 0.75. $mAP0.5:0.95_{x}$ indicates object area in $pixels^2$ where \emph{\textbf{x}} can be \emph{\textbf{s}}, \emph{\textbf{m}} or \emph{\textbf{l}}. \emph{\textbf{s}} = small $\implies area < 32^2$, \emph{\textbf{m}} = medium $\implies 32^2 < area < 96^2$ and \emph{\textbf{l}} = large $\implies area > 96^2$\cite{COCO2024}.
\end{tablenotes}
\end{table}
\subsubsection{Vertical Edges}
 \begin{figure*}[t]
    \centering
    \captionsetup[subfloat]{captionskip=0pt, textfont=sf} 
    \subfloat[\emph{Faster RCNN}\label{fig:faster-rcnn}]{\includegraphics[width=3.3in, keepaspectratio]{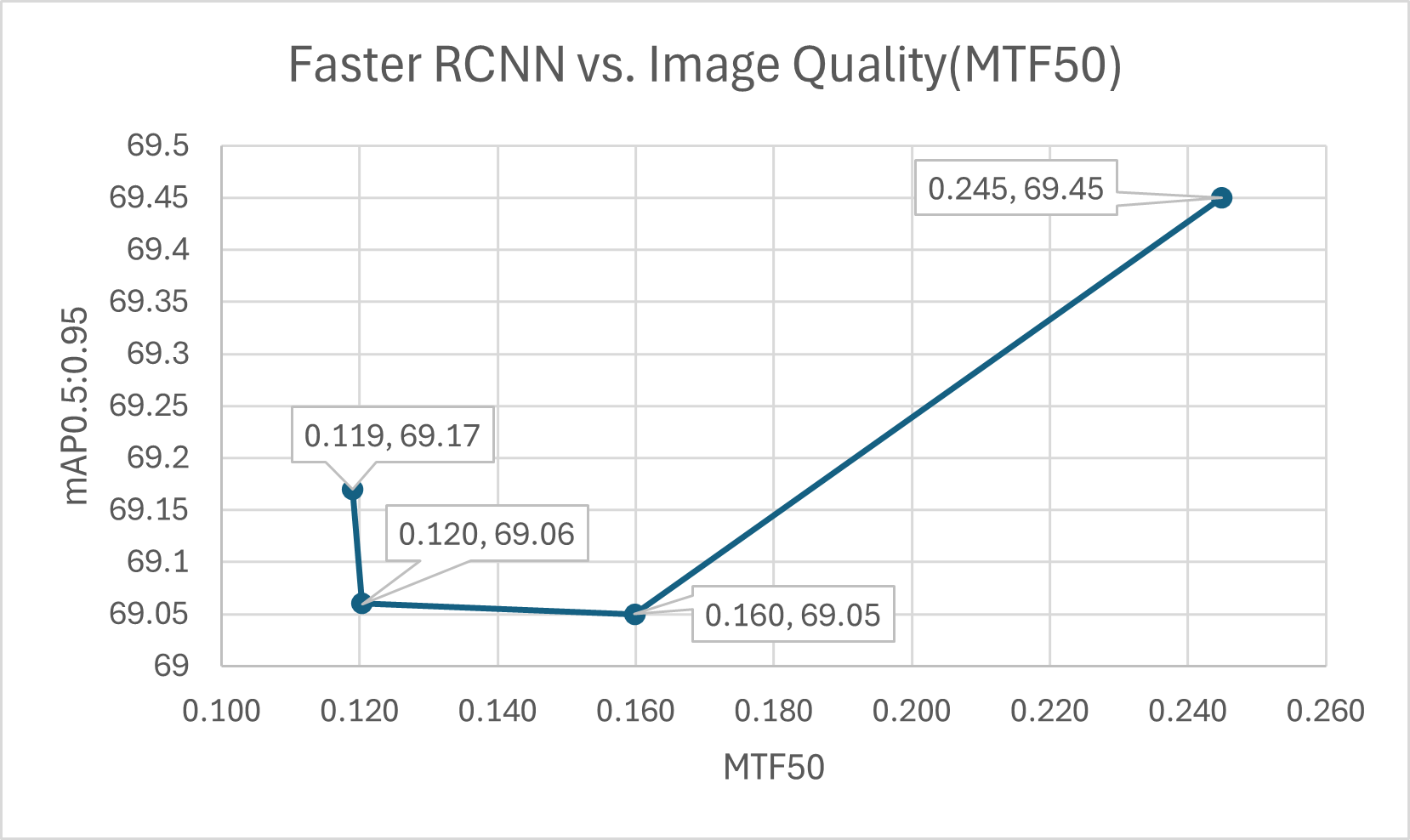}}
    \hspace{0.1cm}
    \subfloat[\emph{YOLOF}\label{fig:yolof}]{\includegraphics[width=3.275in, keepaspectratio]{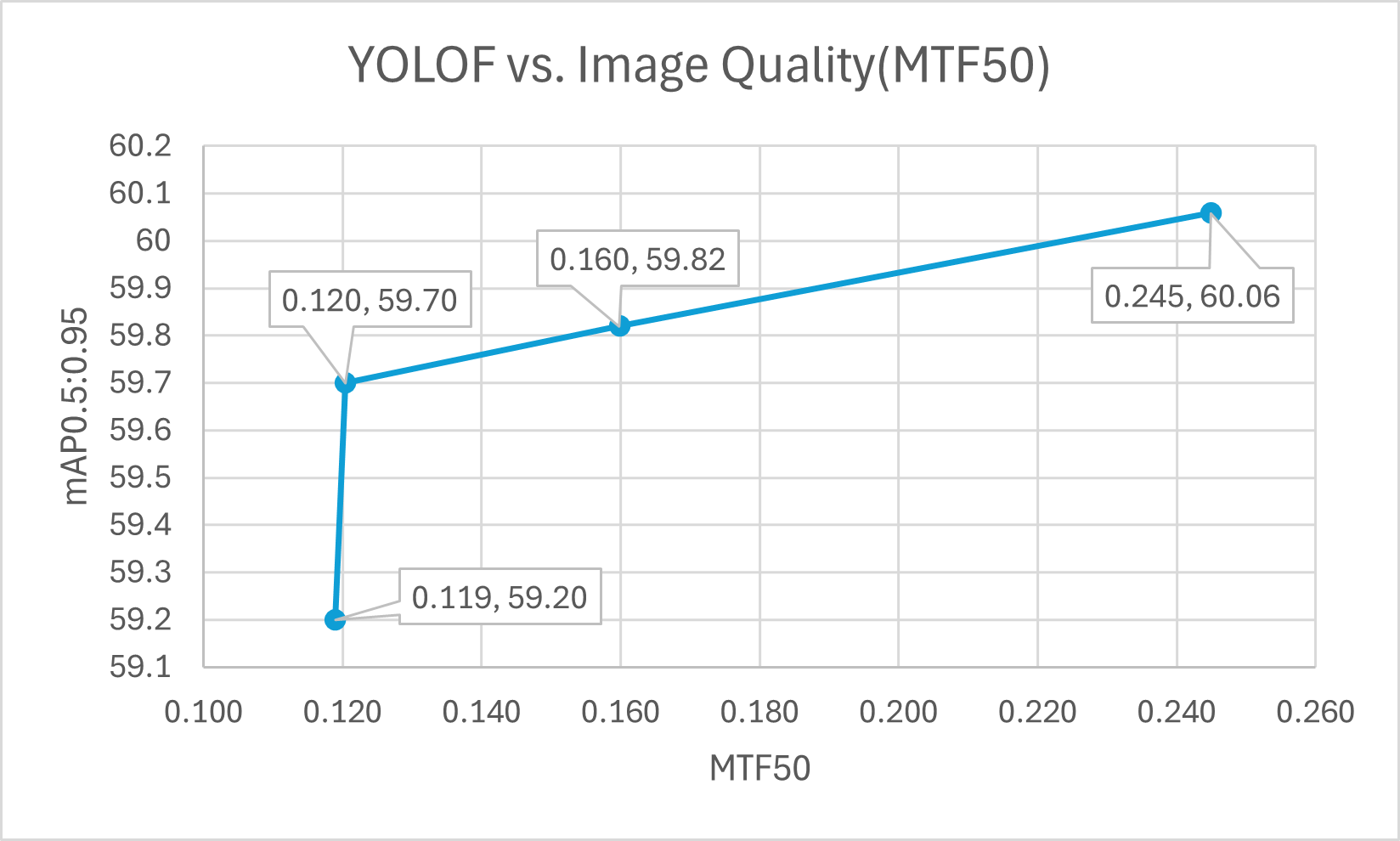}}\\
    \subfloat[\emph{DETR}\label{fig:detr}]{\includegraphics[width=3.3in, keepaspectratio]{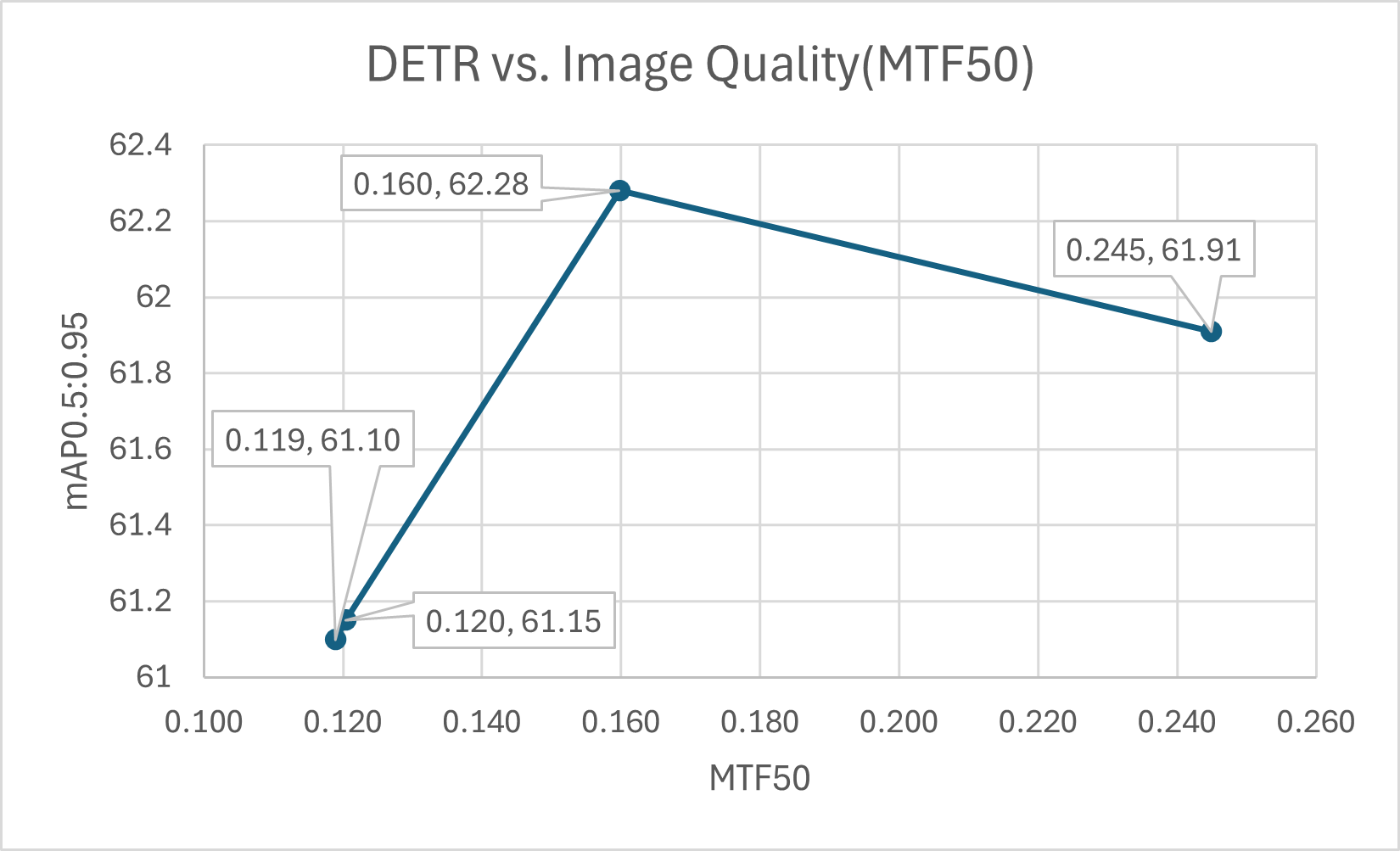}}
    \caption{Mean MTF50 versus mean Average Precision with IoU of between 0.5 and 0.95 (mAP0.5:0.95) for the Virtual KITTI baseline and respective degradations between $\sigma=[1,2,3]$. Mean MTF50 was calculated by averaging results between HMTF50 and VMTF50 from Table \ref{tab:results} creating the following: $MTF50(cy/px)=[0.119,0.120, 0.160, 0.245]$(see Table \ref{tab:results} for \emph{MTF50(mean)}).}
    \label{fig:apvmtf50}
\end{figure*}
Similar to HMTF50, Vertical MTF50(VMTF50) can be seen in Table \ref{tab:results} where four different results can be measured for MTF50. The baseline dataset was measured to be around 0.256cy/px which is slightly higher than for horizontal edges. With $\sigma=1$ and kernel size of $(k_{1},k_{1}) = (7,7)$ MTF50 results degraded by 37.89\%. With $\sigma=2$ and kernel size of $(k_{2},k_{2}) = (13,13)$ MTF50 results degraded by 52.34\%. Finally, with $\sigma=3$ and kernel size of $(k_{3},k_{3}) = (19,19)$ MTF50 results degraded by 53.52\%. Please see Table \ref{tab:results} for VMTF50 results.  $\sigma=3$ generated the largest degradations compared to the baseline.

\subsection{Object Detection}
Object detection results are shown in Table \ref{tab:cv-results}. As can be seen, the Faster RCNN model produced the highest results where overall mAP0.5:0.95 was between 69.05-69.45\%. The YOLOF model produced the lowest results where overall mAP0.5:0.95 was between 59.20-60.06\%. It is clear from Table \ref{tab:cv-results} that the YOLOF model did not perform as well with small objects where mAP0.5:0.95$\textsubscript{s}$ was between 26.16-33.03\%. The Faster RCNN model was also 5.27\% higher than YOLOF for medium objects. After 150 epochs, DETR performance exceeded YOLOF by at least 1.73\% and mAP0.5:0.95 results were between 61.10-62.28\%. DETR also had the lowest results for medium objects where mAP0.5:0.95$\textsubscript{m}$ was between 70.54-73.04\%.

\subsection{Discussion}
While there are very small differences between both $\sigma=2$ and $\sigma=3$, it can be concluded that $\sigma=3$ with $(k_{3}, k_{3})=(19,19)$ provided the largest degradation out of all variations (see Table \ref{tab:results}). Object detection results in Table \ref{tab:cv-results}, show that despite the degradation effects of Gaussian blur, there is no substantial degradation in results and the models are robust. For example, Figure \ref{fig:apvmtf50} shows only small decreases in mAP0.5:0.95 of approximately 0.58\% for Faster RCNN (see Figure \ref{fig:faster-rcnn}), 1.45\% for YOLOF (see Figure \ref{fig:yolof}) and 1.93\% for DETR (see Figure \ref{fig:detr}) across all four variations of the held-out test sets for Virtual KITTI. Faster RCNN is a two-stage object detection architecture that predicts region proposals, allowing it to focus on relevant regions in an image, leading to higher accuracy in detecting small and dense objects in images. This is evident in the significant jump in mAP for small objects(see Table \ref{tab:cv-results}). In contrast, YOLOF uses Uniform Matching to achieve a balance on positive samples with sparse anchors where the architecture bridges the gap in performance with Feature Pyramid Networks (FPN) typically used with Faster RCNN. However, YOLOF struggles to detect small objects closely packed together which is associated with the spatial constraints of the YOLO algorithm. Using the '1x' configuration 
from MMDetection\cite{mmdetection}, YOLOF can produce an overall mAP that is within 1.04 mAP of DETR reflecting the observations of the original paper where a relatively low number of epochs is required to match DETR performance \cite{chen2021you}. These findings might suggest that an out-of-focus camera system, even with considerable image sharpness degradations, may not significantly impact autonomous perception systems for object detection.

\section{Conclusion}
In this paper, we have investigated the impact of Gaussian blur on both image sharpness and object detection performance of three different computer vision models for the Virtual KITTI dataset. This is a crucial step in understanding how image quality degradation might impact a scene in real life. Overall, object detection results show a consistent trend in stable performance where it was found that while image quality degrades from an average of 0.245cy/px to approximately 0.119cy/px; object detection performance stays largely robust where models trained on degraded images can generalize and learn Gaussian blur from the images. However, a slight downward trend can be observed in some cases where there is a decrease in values for the degraded images and the baseline between the lowest and highest values. There are decreases of approximately 0.58\%(Faster RCNN), 1.45\%(YOLOF), and 1.93\%(DETR) across all held-out test sets. Future work that can be considered is the impact of Gaussian blur degradations in other applications such as Depth Estimation or Instance Segmentation for Virtual KITTI. We also recognise that Gaussian blur is a simplistic model of image degradation and other well-known degradations exist such as the defocus blur corruptions of the Hendrycks and Dietterich ImageNet-C dataset \cite{hendrycks2019benchmarking}. In future work, we intend to integrate more realistic models, such that we can examine the impact of realistic camera degradations on computer vision performance. More realistic optical degradation models can be, for example, an optical doublet or a fisheye lens with ray-tracing performed and simulated to degrade images, introducing realistic optical artefacts into simulation.

\section{Acknowledgments} 
This work was supported with the financial support of the Science Foundation Ireland grant 13/RC/2094\_2. The authors of this work would like to thank Dr. Eoin Martino Grua for his invaluable support and encouragement, shaping the direction and the outcome of this study.
\bibliographystyle{apalike}

\bibliography{imvip}

\end{document}